# Review of AlexNet for Medical Image Classification

Wenhao Tang[1], Junding Sun[1,*], Shuihua Wang[2], Yudong Zhang[1,3,†]

[1]1st School of Computer Science and Technology, Henan Polytechnic University, Jiaozuo, Henan 454000, PR China
[2]2nd Department of Biological Sciences, Xi'an Jiaotong-Liverpool University, Suzhou, Jiangsu 215123, China
[3]3rd School of Computer Science and Engineering, Southeast University, Nanjing, Jiangsu 210096, China

## Abstract

In recent years, the rapid development of deep learning has led to a wide range of applications in medical image classification. The variants of neural network models with ever-increasing performance share some commonalities: to try to mitigate overfitting, improve generalization, avoid gradient vanishing and exploding, etc. AlexNet first utilizes the dropout technique to ease overfitting and the ReLU activation function to prevent vanishing gradient. Therefore, we focus on AlexNet, which initially contributed significantly to Convolutional Neural Networks (CNNs) research in 2012. After reviewing over 100 papers, including those from journals and conferences, we give a narrative on the technical details, advantages, and application areas of AlexNet.

## 1. Early Development of Deep Learning

During the development of Deep Learning (DL), artificial neural networks and machine learning techniques have made great strides, leading to the emergence of powerful deep neural networks. The evolvement of DL has gone through four stages:
Early Developments: The concept of artificial neural networks dates back to the 1940s and 1950s, with the development of perceptrons by Frank Rosenblatt [1]. However, early neural networks had limitations, and research stagnated. The perceptron, a single-layer neural network, could only model linearly separable functions, limiting its practical applications.

Revival with Backpropagation: In the 1980s, the field of neural networks experienced a resurgence with the development of the backpropagation (BP) algorithm, which allowed for efficient training of multi-layer neural networks [2]. Researchers such as Geoffrey Hinton and Yann LeCun contributed significantly during this period. However, due to the disappearance of gradients, deep networks still face challenges in training and scaling, where gradients become weaker and weaker as they propagate layer by layer until they eventually disappear, no longer contributing to the learning process [3].

DL Renaissance: The DL Renaissance, occurring primarily in the 2010s, marked a transformative period in the field. Breakthroughs in neural network architectures, increased computational power, the availability of large datasets, and the introduction of deep learning frameworks fueled significant advancements. Section 2 will present detailed content related to the DL Renaissance.

Breakthroughs and Applications: The content related to breakthroughs in DL will be introduced in Section 3. The success of AlexNet demonstrated the power of deep convolutional neural networks (CNNs) for image recognition. In the following years, DL methods have achieved remarkable results in many areas such as natural language processing, speech recognition, robotics, and self-driving cars. This led to the widespread adoption of DL techniques and advanced deep neural network architectures.

*Corresponding author. Email: sunjd@hpu.edu.cn
†Corresponding author. Email: yudongzhang@ieee.org

Our paper is structured as follows: Section 2 focuses on two essential innovations in DL and describes the reasons and conditions that have led to its revival. Then, Section 3 describes how the success of AlexNet has led to breakthroughs in DL. Moreover, in Section 4, we discussed several technologies that AlexNet first used, which improved its performance. Then, in Section 5, we discussed the main operational processes and constituent structures of convolutional layers. The role and some concepts of ReLU activation function and dropout technology in improving model performance are mentioned in Sections 6 and 7, respectively. Next, in Sections 8 and 9, we successively introduce various variants of AlexNet and summarize and discuss the application areas of AlexNet. We will specifically discuss the application of AlexNet in medical diagnosis in Section 10. Finally, Section 11 mentions the conclusions we have drawn based on the specific content of this article.

## 2. Renaissance of Deep Learning

There is resistance to the DL journey. Although the emergence of backpropagation algorithms has brought new opportunities for developing neural networks, the difficulty of training and tuning [4] has gradually increased with the rise of deep networks. Traditional DL methods rely on large amounts of labeled data [5], which comes at a significant cost, which reflects the training challenge. Researchers started exploring different solutions to solve the problem of vanishing gradients. Among them, the proposal of ResNet opens a new path for deep network training. One of Resnet's innovations is skip connection. With $X$ denoting the output of $l$-th layer, $F(X)$ denoting the output of the weighted $(l + 1)$-th layer with skip connection added, the output of $(l + 1)$-th layer is:

$$F'(X) = X + F(X). \quad (1.)$$

By introducing a skip connection (Figure 1), the residual structure breaks the layer-stacking approach of the traditional network so that the gradient can be directly transferred between different layers, thus effectively alleviating the problem of gradient vanishing [6].

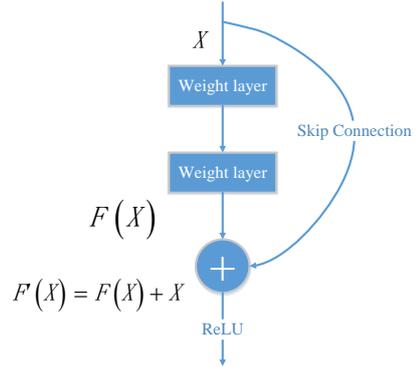

**Figure 1.** Diagram of skip connection.

The Batch Normalization (BN) technique appeared during this period. This technique mitigates the internal covariate bias problem by normalizing the inputs of each layer in the network [7], which speeds up the training of the network and improves the generalization ability of the model. However, it harms the model's robustness against adversarial examples [8]. Kong, et al. [8] further explained the vulnerability of batch normalization in terms of new aspects, such as intra- and inter-class feature distances. BP neural networks affect complex model fitting and distribution approximation that traditional statistical methods cannot achieve. It has excellent potential for development compared to many machine learning algorithms [2]. It has also facilitated many innovations in the field of DL, such as Convolutional Neural Networks (CNNs) and Recurrent Neural Networks (RNN) [9]. These network architectures have achieved remarkable success in many tasks, such as computer vision and natural language processing [9]. Assume that the input of the BN layer is $B = \{x_1 \dots x_m\}$, and $\gamma$ and $\beta$ are learnable parameters. Then

$$y_i = \text{BN}_{\gamma,\beta}(x_i). \quad (2.)$$

The computational process of BN follows the following steps: first,

$$\mu_B = \frac{1}{m}\sum_{i=1}^{m} x_i, \quad (3.)$$

$$\sigma_B^2 = \frac{1}{m}\sum_{i=1}^{m}(x_i - \mu_B)^2, \quad (4.)$$

where $\mu_B$ and $\sigma_B^2$ are the mean and variance of $B$, respectively. Then, normalizing $x_i$ using mean $\mu_B$ and variance $\sigma_B^2$, namely

$$\tilde{x}_i = \frac{x_i - \mu_B}{\sqrt{\sigma_B^2 + \varepsilon}}. \quad (5.)$$

The purpose of normalization is to regularize the data to a uniform interval, reduce the degree of dispersion of the data, and reduce the learning difficulty of the network. Finally, using $\gamma$ and $\beta$ as the reduction parameters to preserve the distribution of the original data to some extent:

$$y_i = \gamma \tilde{x}_i + \beta. \tag{6.}$$

ResNet and BN are two critical innovations for solving problems related to DL. The DL renaissance began in the 2010s when researchers devised techniques to address the challenges of training deep neural networks. Notably, in 2006, Hinton, *et al.* [10] introduced the Restricted Boltzmann Machine (RBM) and deep belief networks (DBNs), paving the way for the resurgence of DL, additionally developing the ReLU activation function, dropout regularization, and advances in hardware (e.g., GPUs) pretty accelerated training deep networks.

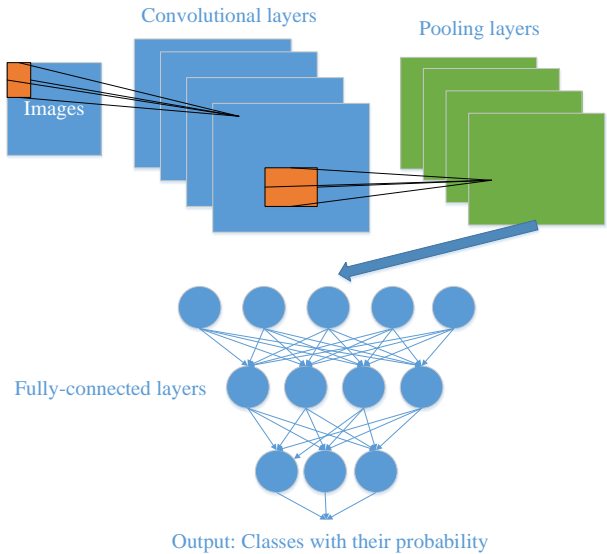

**Figure 2.** The simple architecture of CNN.

Today, DL is a dominant paradigm in artificial intelligence and machine learning, enabling state-of-the-art performance in tasks like image classification (achieved excellent results since 2012 [11]), language translation [5], and autonomous driving [12]. The field continues to evolve, with ongoing research into advanced architectures, transfer learning (TL), reinforcement learning, and the application of DL to new domains [5], promising continued innovation and transformative breakthroughs.

## 3. AlexNet: A Groundbreaking Convolutional Neural Network

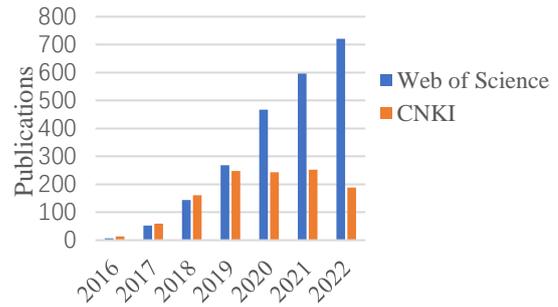

**Figure 3.** Publication record for topic-relevant publications related to AlexNet in Web of Science and CNKI from 2016 to 2022.

The breakthrough moment for DL came in the 2010s, driven by several key developments, including AlexNet's victory in the 2012 ImageNet competition [13], which showcased the power of Deep Convolutional Neural Networks (DCNNs) for image recognition. As shown in Figure 3, there has been a continuous rise in the circulation of publications related to the AlexNet topic through topic-based searches on Web of Science and China National Knowledge Infrastructure (CNKI). This year, the number of topic-relevant publications related to AlexNet exceeds 500 and 200, respectively. We can also see from Table 1 that the years with the highest number of publications about AlexNet on CNKI and Web of Science were 2021 and 2022, which amounted to 252 and 716, respectively. In addition, many related works are not in these two comprehensive paper search sites. The above shows that AlexNet, in DL, is a hot research object.

Table 1. AlexNet-relevant publications from 2016 to 2022.

| Year | Database | |
|---|---|---|
| | Web of Science | CNKI |
| 2016 | 7 | 14 |
| 2017 | 53 | 59 |
| 2018 | 144 | 161 |
| 2019 | 269 | 248 |
| 2020 | 467 | 243 |
| 2021 | 597 | 252 |
| 2022 | 721 | 188 |

AlexNet is an extensive network structure with 60 million parameters and 650,000 neurons [14]. It significantly outperformed traditional methods, drew attention to the potential of DL, and laid the foundation for the subsequent development of Deep Convolutional Neural Network (DCNN). AlexNet represented a significant leap forward in image classification tasks. The network's architecture was a DCNN consisting of eight layers. The eight layers are five convolutional layers and three Fully Connected (FC) layers. The role of the softmax layer is to control the output in the range of (0,1) to ensure neuron activation [14]. Eq. 7. [14] expressed the normalization process of the softmax layer.

$$\text{softmax}(x_i) = \frac{\exp(x_i)}{\sum_{j=1}^{n} \exp(x_j)}, \qquad (7.)$$

where $n$ is the dimension of the input tensor and $x_i$ is also its $i$-th predicted value. One of AlexNet's groundbreaking innovations was using Rectified Linear Units (ReLU) as activation functions, greatly accelerating training convergence. AlexNet also effectively used data augmentation and dropout techniques, which helped prevent overfitting. AlexNet has had considerable significance in previous generations of CNNs and has influenced many subsequent architectures and innovative research in CNNs [15].

## 4. Architecture of AlexNet

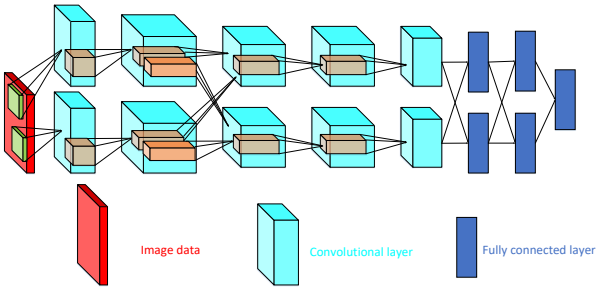

**Figure 4.** Architecture of AlexNet.

As shown in Figure 4, AlexNet [13] consists of five convolutional layers, with the first two being followed by max-pooling layers. Its convolutional layers extract hierarchical features from input images. They employ relatively small filter sizes (e.g., $3 \times 3$ and $5 \times 5$) and utilize many filters (e.g., 96 and 256) in the early layers. These layers are responsible for learning low-level features, such as edges and textures, and gradually building up to more complex representations.

ReLU activation functions are applied after all convolutional layers and the first two FC layers. The main advantage of ReLU is its improved gradient propagation, and it produces fewer problems with gradient vanishing [16]. ReLU introduces non-linearity to the network by setting negative values to zero while allowing positive values to pass unchanged [17]. This non-linearity promotes learning complex and hierarchical features and accelerates training by mitigating the vanishing gradient problem. Moreover, in neural networks, activation functions add nonlinear properties so that the network can learn and perform more complex tasks. ReLU, although widely used, still suffers from the computational complexity of training multi-layer fully connected ReLU neural networks [18].

AlexNet introduced Local Response Normalization (LRN) to enhance the network's generalization ability. LRN normalizes the responses of neighboring neurons, promoting competition between neurons and improving the network's robustness to variations in input data. The following is the formula for LRN:

$$b_{x,y}^i = \frac{a_{x,y}^i}{\left(k + \alpha \sum_{j=\max\left(0, i-\frac{n}{2}\right)}^{\min\left(N-1, i+\frac{n}{2}\right)} \left(a_{x,y}^j\right)^2\right)^\beta}, \qquad (8.)$$

where $i$ denotes the $i$-th channel, $x$ and $y$ denotes the positional coordinates of this value to be normalized, $k$ serves to prevent the occurrence of division by 0, $\alpha$ and $\beta$ are constants; $n$ denotes the range of the neighborhood, with the boundary cases complemented by 0.

However, BN [19] techniques emerged later to improve higher performance than LRN. The core idea of BN is to constrain the change in gradient to a transferable range, preventing the gradient from vanishing or exploding. BN attempts to achieve a stable distribution of activation values throughout the training process by subtracting the batch mean and dividing by the batch standard deviation [20]. Jain, et al. [21] proposed a new high-performance convolutional Siamese network discussing the problem of offline signature verification. In this model, they used the BN technique instead of local response normalization to get better accuracy.

For AlexNet, three max-pooling layers follow the first three convolutional layers. These pooling layers downsample the spatial dimensions of the feature maps, reducing computational complexity while retaining essential information. Max-pooling selects the maximum value within a local region to create a downsampled representation. After the convolutional and pooling layers, AlexNet incorporates three FC layers. These layers are traditional artificial neural network layers where all neurons in every layer have links with every neuron in the next layer. A softmax activation function is behind the final FC layer, which produces class probabilities, making it suitable for image classification tasks.

AlexNet introduced dropout as a regularization technique to prevent overfitting. Dropout randomly deactivates a fraction of neurons during training, reducing co-dependencies between neurons and improving model generalization. AlexNet was designed to take advantage of parallel processing capabilities and was one of the first CNNs to do so effectively. It utilized two GPUs during training, which allowed it to handle the large number of parameters and computations involved in DL.

## 5. Convolutional Layers

The primary operation performed in convolutional layers is convolution. Convolution involves sliding a set of learnable filters (kernels) over the input data. These filters are small grids that have weights associated with each element. As shown in Figure 5, the convolution operation calculates the dot product between the filter and the local region of the input data.

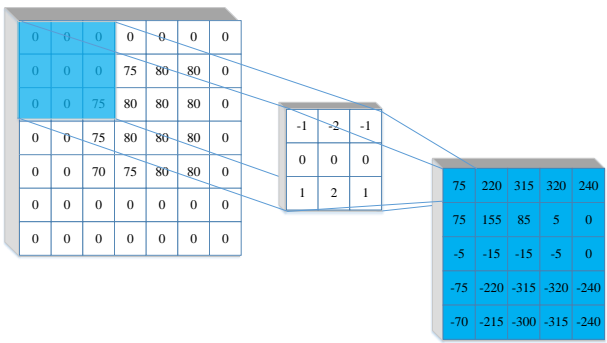

**Figure 5.** A simple graphical representation of the convolution operation.

This operation is repeated across the entire input to produce feature maps. In CNNs, the filters are learnable parameters updated during training through backpropagation. Each filter detects specific patterns or features in the input data. For example, some filters may specialize in detecting edges, while others may examine textures, corners, or other visual features. As the network learns, filters become increasingly specialized and optimized for the given task. Convolutional layers have a hierarchical organization. Early layers focus on extracting low-level features from the input data, such as edges and simple textures. These features serve as building blocks for more complex representations. As the layers progress, they learn higher-level features and patterns, eventually capturing abstract concepts related to the task. The feature map of the original size $n$ will probably change after it passes through the convolutional layer with kernel size $k$. The following equation describes the process of 2D convolution:

$$W = \left\lfloor \frac{n-k+2p}{s} \right\rfloor + 1, \qquad (9.)$$

where $W$ is the size of the 2D post-convolution feature map, $k$ is the size of the convolution kernel, $s$ is the pooling unit move step, $p$ is the number of laps of the expanded pixels.

Often, after the convolution operation, a pooling layer and an activation function (usually ReLU) follow. Pooling layers, such as max-pooling, downsample the spatial dimensions of the feature maps while retaining important information. Activation functions introduce non-linearity to the network, allowing it to learn complex relationships. ReLU, in particular, is widely used in convolutional layers to activate positive values and set negative values to zero, promoting sparsity and faster convergence.

## 6. ReLU Function

The ReLU activation function is fundamental to specific artificial neural networks. Deep neural networks are a representation of them. It plays a crucial role in introducing non-linearity to neural networks, enabling them to learn complex relationships in data. Here's an overview of the ReLU activation function in four paragraphs: ReLU is a simple mathematical function that operates on an input value $x$. As shown in Figure 6, it returns $x$ if $x$ is greater than or equal to zero and zero otherwise. The formula can describe the above:

$$y = \begin{cases} x, & x \in [0, +\infty) \\ 0, & x \in (-\infty, 0) \end{cases}, \qquad (10.)$$

where $y$ and $x$ are the output and input of ReLU, respectively. Eq. 10. denotes that ReLU replaces all negative input values with zero and leaves positive values unchanged. This piecewise linear behavior introduces non-linearity into the network, allowing it to approximate complex functions and learn intricate patterns.

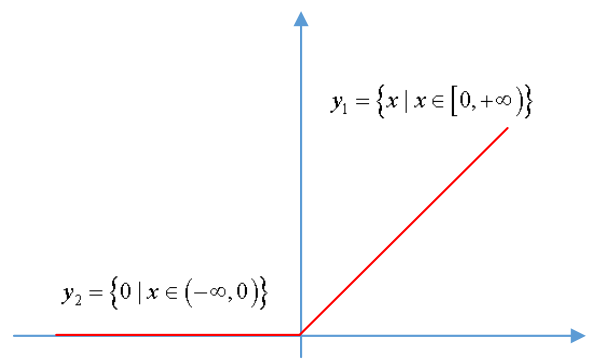

**Figure 6.** Graphical representation of ReLU.

ReLU has several advantages, making it a popular choice as an activation function in neural networks. Firstly, it is computationally efficient because it involves a simple thresholding operation. Secondly, ReLU helps alleviate the vanishing gradient problem, which can hinder the training of deep networks. In traditional activation functions like sigmoid and tanh, gradients can become very small in the presence of deep architectures, slowing down or preventing learning. ReLU mitigates this problem by maintaining a constant gradient for positive input values, ensuring faster convergence during training. Lastly, ReLU activation sparsity—setting negative values to zero—can lead to more efficient representations, encouraging the network to focus on relevant features.

Over the years, several variants and modifications of ReLU appeared to address its limitations. One such variant is the Leaky ReLU, which allows a slight, non-zero gradient for negative input values to address the "dying ReLU" problem, where neurons can get stuck with zero gradients and cease to update. Another variant is the Parametric ReLU (PReLU) [22], which introduces a learnable parameter $\alpha$ to adjust the slope of the negative side of the function. This can be particularly useful when the optimal slope may vary across different neurons and layers. And the following is the equation of PReLU:

$$f_P(x) = \max(0, x) + \alpha \cdot \min(0, x), \alpha \in [0,1], \quad (11.)$$

where, $f_P(x)$ is the output of the PReLU function, $f(x) \in (0, +\infty)$, $x$ is its input.

There are also variants like Exponential Linear Unit (ELU) [23] and Swish [24], each designed to offer different advantages in terms of convergence and robustness. For the ELU function, the output $f_E(x)$ grows linearly for input values $x$ not less than 0. Conversely, $f_E(x)$ decreases exponentially. This property makes ELU better than ReLU, especially when dealing with negative inputs. The ELU function has the following formula:

$$f_E(x) = \begin{cases} x & , x \geq 0 \\ e^x - 1 & , x < 0 \end{cases}. \quad (12.)$$

The main advantage of the Swish function is that it has a high-speed convergence rate and effectively avoids the problem of vanishing and exploding gradients. The formula for the Swish function is as follows:

$$f_S(x) = x * \text{sigmoid}(x), \quad (13.)$$

where sigmoid and $*$ denotes the Sigmoid function and the element-wise multiplication operation, respectively. $f_S(x)$ and $x$ are the output and input of the Swish function, respectively.

ReLU is an activation function in various neural network architectures, including CNNs for computer vision tasks, recurrent networks for natural language processing, and fully connected networks for multiple applications. It has been a critical factor in the success of DL, enabling networks to learn complex representations and achieve state-of-the-art results in tasks like image classification, object detection, and more. While ReLU is highly effective, choosing the appropriate activation function sometimes depends on the specific problem and architectural considerations. Researchers continue to explore alternatives and improvements in activation functions to address various challenges in DL.

## 7. Dropout

As shown in Figure 7, dropout works by randomly deactivating a fraction of neurons during each training iteration [25]. But dropout cannot be helpful when just a single hidden node [25]. During forward and backward passes, dropout randomly sets a subset of neuron activations to zero with a specified probability, typically between 0.2 and 0.5. The process means some neurons are "dropped out" or temporarily removed from the network for that specific iteration. Dropout introduces noise and uncertainty into the training process, forcing the network to rely on different neurons in each iteration.

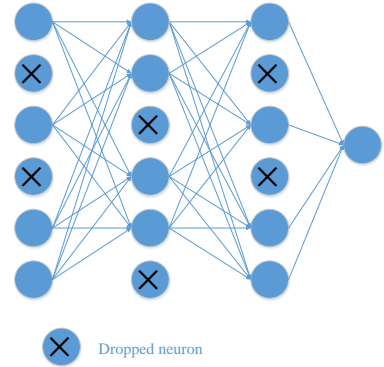

**Figure 7.** Simple graphical representation of dropout. The figure explicitly shows that dropped neurons have lost connections to neurons in neighboring layers.

Overfitting occurs when a neural network learns to memorize the training data rather than generalize. It leads to poor performance on unseen data. Dropout combats overfitting by preventing co-adaptation of neurons [26]. When randomly dropping out neurons during training, the network cannot rely on any single neuron or small group of neurons to make predictions. Instead, it must distribute the learning across a more diverse set of neurons, encouraging the network to learn robust and generalizable features.

Dropout is an ensemble learning technique to some extent. Multiple subnetworks with different neuron configurations are sampled from the whole network during training. By aggregating the predictions of these subnetworks during inference (usually by scaling the activations by the dropout probability at test time), dropout effectively creates an ensemble of networks. Ensemble learning tends to reduce model variance and improve overall performance.

Variations of dropout have been proposed, including dropout applied to different network layers, such as dropout in convolutional layers (DropBlock) [27], and more. These variants adapt dropout to specific network architectures and tasks. Dropout is a powerful and widely used regularization technique that helps neural networks generalize better to unseen data, making them more robust and less prone to overfitting [28]. For a linear neural network without dropout, its error $E_N$ can be expressed as:

$$E_N = \frac{1}{2}\left(t - \sum_{i=1}^{n} w_i I_i\right)^2, \quad (14.)$$

where $w$ is the set of weights of the neural network layer, $t$ is the target value, $I$ is the set of inputs to the neural network layer. After adding the dropout layer, the error will change as the following:

$$\begin{cases} E'_N = \frac{1}{2}\left(t - \sum_{i=1}^{n} w'_i I_i\right)^2, \\ w' = pw \end{cases} \quad (15.)$$

where $E'_N$ is the changed error, and $p$ is the dropout rate. Dropout has been instrumental in the success of DL in various domains, from computer vision to natural language processing, and remains a valuable tool for improving model performance and reliability.

## 8. Variants of AlexNet

The classical structure of AlexNet laid the foundation for the field of DL, while some researchers have made some modifications and improvements to it, resulting in different variants. Some common variants of AlexNet, for example, the earlier appearing ZFNet, are improvements on AlexNet proposed by Zeiler*, et al.* [29]. It builds on AlexNet with some tweaks aimed at improving performance and a better understanding of the hierarchical structure of images. Subsequently, Simonyan*, et al.* [30] proposed VGGnet with a deeper network structure. Compared to AlexNet, VGGNet has a characteristic that uses smaller convolutional kernels and deeper networks to increase the depth of the network. In the same year GoogLeNet, also known as Inception v1, was proposed by Szegedy*, et al.* [31]. It introduced the "Inception module" to increase the computational efficiency of the network by using different-sized convolutional kernels in parallel. Later, Microsoft Research proposed ResNet, which made it easier to train deep structures by introducing a residual learning mechanism. ResNet was designed to solve the problem of gradient vanishing and gradient explosion during deep neural network training. Also, SqueezeNet [32] is a lightweight neural network designed to reduce the model's size while maintaining good performance. It reduces the number of parameters by employing a 1×1 convolutional kernel, making it suitable for resource-limited environments. The later proposed DenseNet is a DL architecture proposed by Huang*, et al.* [33]. Compared to traditional CNNs, DenseNet utilizes a densely connected structure, which connects each layer's output to the outputs of all previous layers. This densely connected design helps to enhance gradient flow, facilitates information sharing, and effectively mitigates the gradient vanishing problem.

All of these networks are based on CNN and contain convolutional layers, pooling layers, etc.; all of them are deeper than earlier convolutional networks and, therefore, more effective in extracting features; all of them have been trained and validated on large-scale image datasets, and all of them have better performance. These networks differ in network structure, parameter utilization, depth, and application scenarios.

## 9. Applications

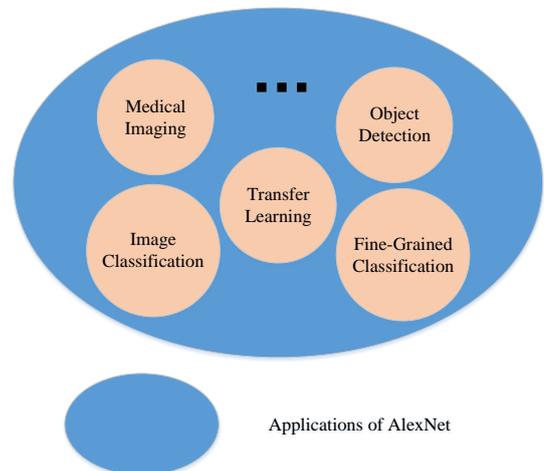

**Figure 8.** A schematic diagram of the application (not fully listed) of AlexNet.

As shown in Figure 8, we will know the application aspects of AlexNet, which include medical imaging, image classification, TL, object detection, and fine-grained classification. The following account addresses these aspects.

Image Classification: AlexNet's primary and most well-known application is image classification. It demonstrated remarkable performance in the ImageNet Large Scale Visual Recognition Challenge (ILSVRC) in 2012 [34], significantly outperforming previous methods. This success established the effectiveness of deep CNNs for image recognition tasks. Since then, similar architectures have broadly applied to various image classification tasks, including identifying objects in images, recognizing faces, and classifying medical images. For facial recognition, for example, Ghazal, et al. [35] developed an innovative transfer learning AlexNet model to detect the presence or absence of autism through children's facial features. This model extracts image features using a convolutional layer of pre-trained AlexNet. It uses these features to train a new FC layer. Comparison experiments on a dataset of 2940 images of children with autism from Kaggle show that their model has higher accuracy. Further, Balashanmugam, et al. [36] applied AlexNet to classify iris images of the eye, achieving a high correct recognition rate on 163,432 iris images, providing valuable insights into the development of biometrics. Kayadibi, et al. [37] applied pre-trained AlexNet to eye condition detection, and their experimental results outperformed the traditional pre-trained deep graph convolutional network.

Object Detection: AlexNet's architecture, particularly its convolutional layers, has been the foundation for developing object detection models. Researchers have adapted and extended the principles of AlexNet to create object detection frameworks like Fast R-CNN [38] and Faster R-CNN [39], which can classify objects and localize them within images. These advancements have applications in autonomous driving, surveillance, natural disaster warning [40], robotics [41, 42], etc.

Fine-Grained Classification: Fine-grained classification involves distinguishing between similar categories within a broader class. AlexNet's capability to learn intricate features has made it suitable for fine-grained classification tasks, such as identifying specific species of birds [43], plant varieties [44, 45], or classification of canine maturity and bone fracture time [46]. Researchers have built upon the principles of AlexNet to develop specialized models for these tasks.

Medical Imaging: AlexNet's CNN architecture was related to medical imaging for various diseases. It has been used for the classification of medical images, such as X-rays, MRIs, CT scans, and electroencephalograms, aiding in the diagnosis of diseases, including cancer [47], neurological disorders [48], and schizophrenia [49]. By adapting AlexNet to medical data, researchers and healthcare professionals have improved accuracy and efficiency in disease detection and diagnosis.

TL: One of the significant applications of AlexNet is in TL. Researchers and practitioners leverage pre-trained AlexNet models, fine-tuning them for specific tasks or domains [50, 51]. This approach efficiently uses pre-trained feature extractors to solve new and related problems with smaller datasets. Sharma, et al. [52] demonstrated experimentally that the hierarchical fine-tuning method has potential for development. TL with AlexNet has enabled advances in various fields, including natural language processing [53, 54], where researchers used the network's features to extract meaningful representations from text data.

In video analytics, the CNN structure of AlexNet has broad applications for dynamic scene understanding. For video classification tasks, researchers use AlexNet to extract critical features in videos for efficient action recognition and event classification. For example, Peng, et al. [55] added additional convolutional and pooling layers to AlexNet and tuned the parameters to learn subpixel-level skeletal image features generated in a certain way. They carefully fine-tuned AlexNet, and their experimental results on the dataset outperformed even VGG16 and GoogLeNet. As another example, Rafiq, et al. [56] used a pre-trained AlexNet and added three additional fully-connected layers to it, applying it to a video scene classification task. They found that batch normalization after the convolutional layers did not improve model performance, so they did not use this technique. Their improvement is relatively simple, although it capitalizes on the benefits of the transfer learning approach and achieves better results on smaller datasets. However, for classification tasks with more video duration and frames, the pre-trained AlexNet, with fewer parameters, may be challenging to handle when more improvements are inexistent. However, simply adding or subtracting layers or tuning parameters to the original model may not improve the model's generalization over larger datasets.

Further, Minhas, et al. [57] applied a DL scheme containing AlexNet to classify sports video footage. Through experiments, they proved that their method performs better than some machine learning methods and standard CNNs. The experiments they did were more detailed and specialized. However, since there is no significant improvement to AlexNet, it may be difficult for their method to perform more complex tasks. In addition, AlexNet's spatio-temporal feature learning capability makes it perform well in video behavior analysis. These applications can support application scenarios such as video surveillance and targeted placement of video advertisements. For example, Mumtaz, et al. [58] proposed a novel Deep Multi-Network (DMN) architecture based on AlexNet and GoogLeNet for video violence detection. They integrated two pre-trained networks to build a fast learning system with high video classification accuracy. Their experiment results demonstrated that the DMN performs better in optimal accuracy and trains faster than AlexNet and GoogleNet. However, DMN is not as good as AlexNet and GoogleNet in some result metrics. The shortcoming may indicate that DMN suffers from information loss or an increased risk of overfitting in fusing two CNN outputs. Secondly, the MobileNet model proposed by Irfanullah*, et al.* [59] performs excellently on the field hockey game dataset. They used the Background subtraction method in the preprocessing stage. This method subtracts the current frame from the background reference model to compute the motion region efficiently.

Further, Khan*, et al.* [60] used a fine-tuned AlexNet for crowd anomaly detection in video frames. This fine-tuned AlexNet with one less convolutional layer extracts features on three anomaly detection public datasets and inputs them to six classifiers for comparative analysis. Among them, the softmax classifier has better performance. AlexNet also has expanded applications to video coding. For example, Imen*, et al.* [61] combined LeNet-5 and AlexNet. Their approach reduces the computational time used to examine block decision candidates. The modifications they made to LeNet-5 were to avoid gradient vanishing and to accommodate multi-label classification, while AlexNet did not change its parameters but downsampled its inputs. Their method reduces the complexity of the HEVC encoder. It also reduces the video processing time and improves the compression efficiency more than related work.

AlexNet uses real-time object recognition and tracking in virtual and augmented reality applications. By combining camera data and features extracted by AlexNet, virtual elements can interact more accurately with the actual scene. For example, the approach proposed by Hung*, et al.* [62] provides physicians with more effective machine-assisted treatment options. They tackle the challenge of capturing the most relevant visual details for better video evaluation through assisted supervision and attention mechanisms.

Although AlexNet is primarily designed for image classification, transfer learning opens up new possibilities for speech recognition tasks. By utilizing the high-level features learned by AlexNet in image recognition, researchers can improve the performance of speech recognition systems and increase the understanding and classification accuracy of speech signals. For example, Ziafat*, et al.* [63] fine-tuned a pre-trained AlexNet by removing the last three layers and did comparison experiments with DCNN and bidirectional Long Short-term Memory (LSTM). The experimental results show that AlexNet, with transfer learning, outperforms the other networks.

AlexNet's fine-grained classification capability in agriculture uses crop yield prediction, plant stress detection, weed and pest detection, disease detection, smart agriculture [64], and sugarcane quality detection [65]. By deploying image acquisition devices in farmland, AlexNet can recognize the health condition of plants [66] and help farmers take timely measures to improve the yield and quality of crops. Zhang*, et al.* [67] developed a targeted shake-and-catch apple harvesting technique to cope with the high cost of manual apple picking. Tarek*, et al.* [68] combined pre-trained AlexNet and an improved Gray Wolf optimization method for early diagnosis of plant diseases. Their model achieved better results on relevant datasets.

AlexNet and its derived models are critical in natural disaster monitoring [69, 70]. By analyzing satellite images or data captured by drones, AlexNet can support disaster assessment [71] and emergency relief decisions [72]. The above shows that it has great potential for mitigating the impact of disasters and improving response capabilities.

## 10. Medical Image Classification Applications

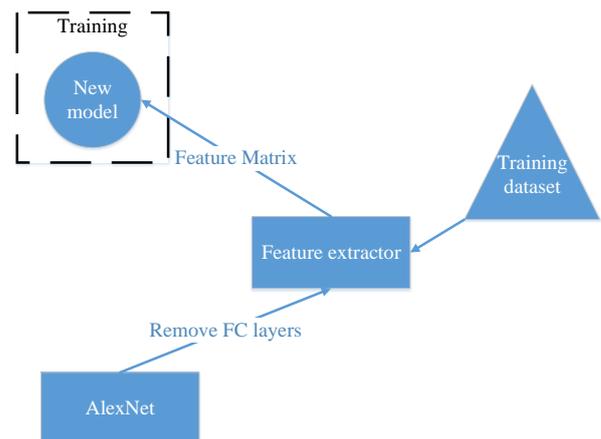

**Figure 9.** A simple illustration of using AlexNet as a feature extractor using TL.

At present, DL has broad applications in the field of medical disease diagnosis. AlexNet has applications in several tasks, but not just these. In these following applications, AlexNet is a feature extractor in a TL (Figure 9) manner, and the extracted features are the input of new models.

Disease Diagnosis: AlexNet is used to diagnose various medical conditions based on imaging data. For example, it has been used in the analysis of radiological images such as X-rays, Magnetic Resonance Imaging (MRI) [73], and Computed Tomography (CT) scans to detect diseases like diseases of the abdomen [74], diseases of the chest [74], diseases of the sexual organs [74], breast cancer [75]. By training on large datasets of medical images, AlexNet can learn to identify patterns and abnormalities indicative of specific diseases,

Wenhao Tang, Junding Sun, Shuihua Wang and Yudong Zhangassisting radiologists and clinicians in making more accurate diagnoses. After a survey of 102 medical texts, Morid, *et al.* [76] stated that AlexNet is the most common model for brain MRI and mammography. AlexNet is also essential to the deep models applied to diagnose brain tumors. The fine-tuned AlexNet has produced good brain tumor dataset results [77]. However, AlexNet also has limitations in diagnosing brain tumors. For example, Ramya, *et al.* [78] proposed MIDNet18 CNN, which is more profound than AlexNet, and experimentally proved that this model performs better than AlexNet on specific brain tumor image datasets. Deeper layers allow the CNN to extract richer features, and AlexNet does not serve as well as deeper CNNs on some datasets due to its shallow layers. AlexNet can be used as a feature extractor with several well-known machine-learning classifiers to improve brain tumor image classification [79].

Lesion Detection and Segmentation: AlexNet uses lesion detection and segmentation in medical images. For instance, it can assist in identifying and delineating skin lesions, including moles, tumors [80], and other dermatological conditions in dermatology. Similarly, in the context of medical imaging, AlexNet can aid in identifying and delineating lesions or anomalies within organs [81-83], tissues [84], or anatomical structures [85], which is essential for treatment planning and disease monitoring.

Ophthalmology: AlexNet has shown promise in ophthalmology, particularly in diagnosing and monitoring eye diseases [86]. It can analyze retinal images, such as fundus photographs [87], to detect signs of diabetic retinopathy [88], glaucoma [89], and age-related macular degeneration [90]. By automatically identifying abnormalities and disease-related features in these images, AlexNet assists ophthalmologists in diagnosing and tracking disease progression early.

Pathology and Histology: In pathology and histology, AlexNet can be applied to the analysis of tissue slides, assisting pathologists in diagnosing diseases like cancer [91]. It can help identify and classify cell types, tissue structures, and pathological conditions in biopsy samples. Automating this process through DL models like AlexNet can improve diagnostic accuracy and reduce the workload of pathologists. For example, Abbas, *et al.* [92] proposed a deep-learning model for recognizing eye diseases using self-attention and dense layers. This model was a variant improved from AlexNet, with the main improvement being the addition of attention and dense layers. They used effective image enhancement techniques such as CLAHE and Bengram. These ways effectively preprocess the dataset. The methods combined with the improved pre-trained AlexNet proved to have high overall efficiency and accuracy, as demonstrated by the experimental results on the ocular disease dataset.

Further, Chan, *et al.* [93] extracted image features with pre-trained AlexNet, VGG16, and GoogleNet to classify ocular diseases [93]. They fused the last FC layer of AlexNet, the first FC layer of Vgg16, and the FC layer of GoogleNet to be used as a decision model. They also reduced the feature space using majority voting. The manipulation ignores some interrelated features and retains the most crucial information. However, this method also results in some data loss.

Overall, the application of AlexNet to medical imaging tasks has the potential to enhance the accuracy and efficiency of disease diagnosis, lesion detection, and image analysis in various medical specialties. Based on the fact that most of the studies on MRI images of the brain and mammography images using AlexNet as a DL model have achieved better results, Morid, *et al.* [76] argued that the best way to conduct such studies is to use shallow CNN models with large convolutional kernel sizes. As DL techniques continue to advance, or with the help of expanding datasets, improved GPU performance, data preprocessing techniques, and growing medical image types to further enhance the performance of CNNs [94], all of these applications are likely to play an increasingly important role in improving medical outcomes and patient care.

## 11. Conclusions

In conclusion, AlexNet, with its pioneering DCNN architecture, has made significant strides in medical image classification. AlexNet models are very effective for faster model training and better feature extraction [95]. Thus, it has opened up new possibilities in disease diagnosis, lesion detection, and image analysis across various medical specialties. The applications of AlexNet in medical imaging are wide-ranging and impactful [77]. It has been successfully employed in the diagnosis of diseases such as brain tumors [77], neurological disorders [96], and ophthalmic conditions [97]. By analyzing MRI scan images [98], radiological images [99], histological samples, and dermatological images, AlexNet assists healthcare professionals in making more accurate and timely diagnoses.

Moreover, AlexNet's contribution to medical imaging extends beyond diagnosis. It has facilitated the automation [100, 101] of tasks such as lesion detection, segmentation, and identifying disease-related features within images. Kora, *et al.* [74] stated that AlexNet and others are the most widely used TL models for medical image analysis. Their word indicates that AlexNet improves diagnostic accuracy and streamlines healthcare practitioners' workflow, leading to more efficient patient care.

As the field of DL continues to advance, and with the development of specialized architectures and the availability of larger medical imaging datasets, the impact of models like AlexNet on medical image classification is expected to grow. These advancements promise to enhance healthcare outcomes, enabling earlier disease detection and ultimately improving the well-being of patients. However, ensuring these AI-powered tools are integrated into clinical practice and collaborating with medical professionals is crucial to achieving the best possible patient care.

## Acknowledgment

We thank all the anonymous reviewers for their hard reviewing work.